%% file: main.tex
\documentclass[runningheads]{llncs}
\usepackage{graphicx}
\usepackage{booktabs} 
\usepackage{amsmath}
\usepackage{algorithm}
\usepackage{algpseudocode}
\usepackage{url}
\usepackage{color}
\usepackage[misc]{ifsym}

\usepackage[belowskip=0pt,aboveskip=0pt]{caption}

\begin{document}
\title{DeepStyle: User Style Embedding for Authorship Attribution of Short Texts}
\titlerunning{DeepStyle}
%
\author{Zhiqiang Hu\inst{1} \and
Roy Ka-Wei Lee\inst{2\footnotemark[1]} \and
Lei Wang\inst{3} \and
Ee-peng Lim \inst{3} \and
Bo Dai \inst{1}}
%
\authorrunning{Hu et al.}
%
\institute{University of Electronic Science and Technology of China, China \and
University of Saskatchewan, Canada \and
Singapore Management University, Singapore \\
\email{zhiqianghu@std.uestc.edu.cn}, \email{roylee@cs.usask.ca}, \email{lei.wang.2019@phdcs.smu.edu.sg}, \email{eplim@smu.edu.sg}, \email{daibo@uestc.edu.cn}}
\maketitle              

\renewcommand{\thefootnote}{\fnsymbol{footnote}} 
\footnotetext[1]{Corresponding author} 

\begin{abstract}
Authorship attribution (AA), which is the task of finding the owner of a given text, is an important and widely studied research topic with many applications. Recent works have shown that deep learning methods could achieve significant accuracy improvement for the AA task. Nevertheless, most of these proposed methods represent user posts using a single type of features (e.g., word bi-grams) and adopt a text classification approach to address the task. Furthermore, these methods offer very limited explainability of the AA results. In this paper, we address these limitations by proposing \textsf{DeepStyle}, a novel embedding-based framework that learns the representations of users' salient writing styles. We conduct extensive experiments on two real-world datasets from Twitter and Weibo. Our experiment results show that \textsf{DeepStyle} outperforms the state-of-the-art baselines on the AA task. 

\end{abstract}

\section{Introduction}

\input{introduction.tex}

\section{Related Works}
\input{related.tex}

\section{Proposed Model}
\input{model.tex}

\section{Experimental Evaluation}
\input{experiment.tex}
\section{Conclusion}
\input{conclusion.tex}

%
%
%
\bibliographystyle{splncs04}
\bibliography{bibliography}

\end{document}


\title{Supplementary}
%
%
%
\author{Anonymous}
%
%
\maketitle              

\section{User Style Embeddings Analysis}
\begin{figure*}[h] 
	\centering
	\setlength{\tabcolsep}{0pt} 
	\renewcommand{\arraystretch}{0} 
	\begin{tabular}{cccc}
		\includegraphics[scale = 0.16]{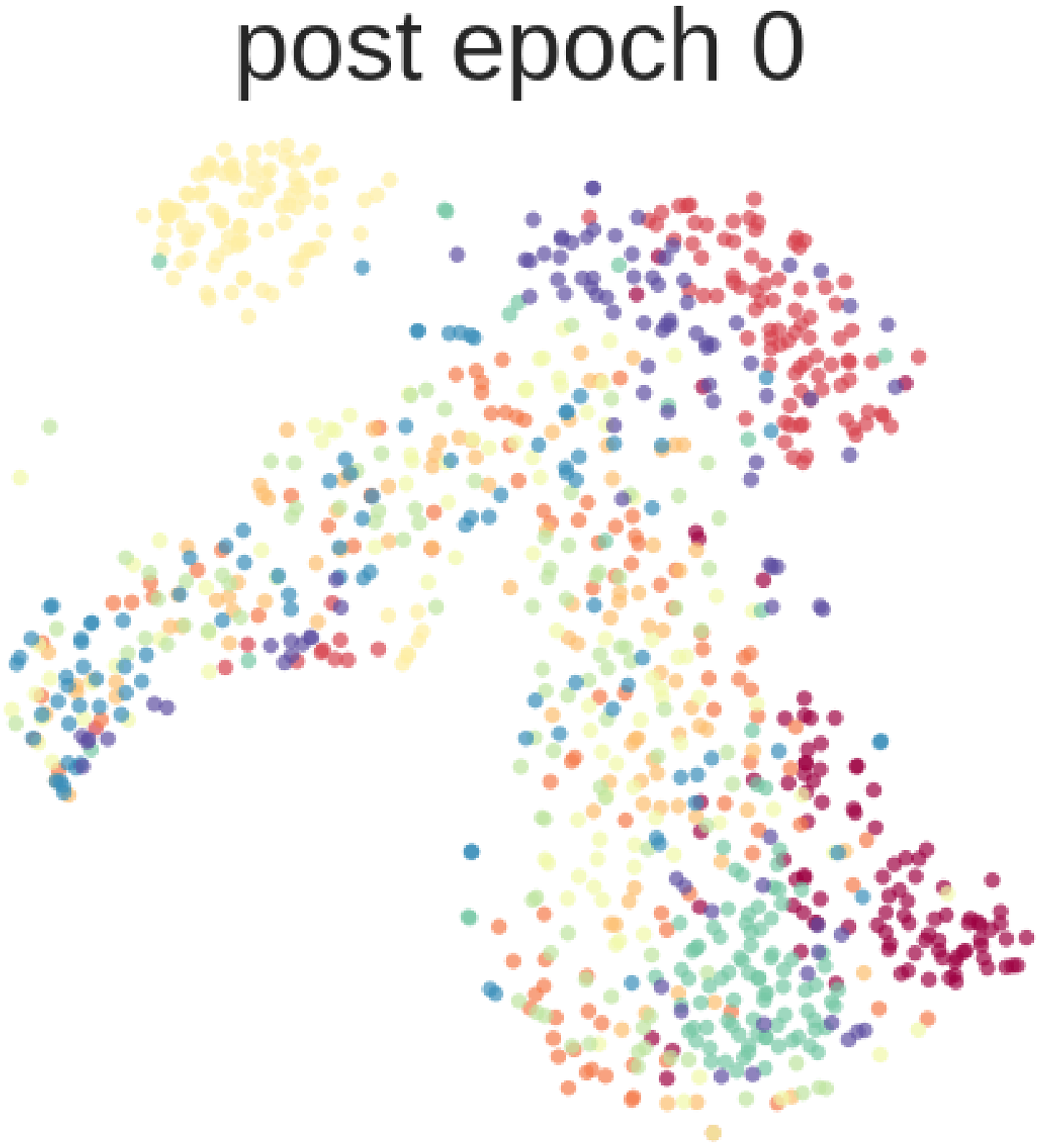} &   
        \includegraphics[scale = 0.16]{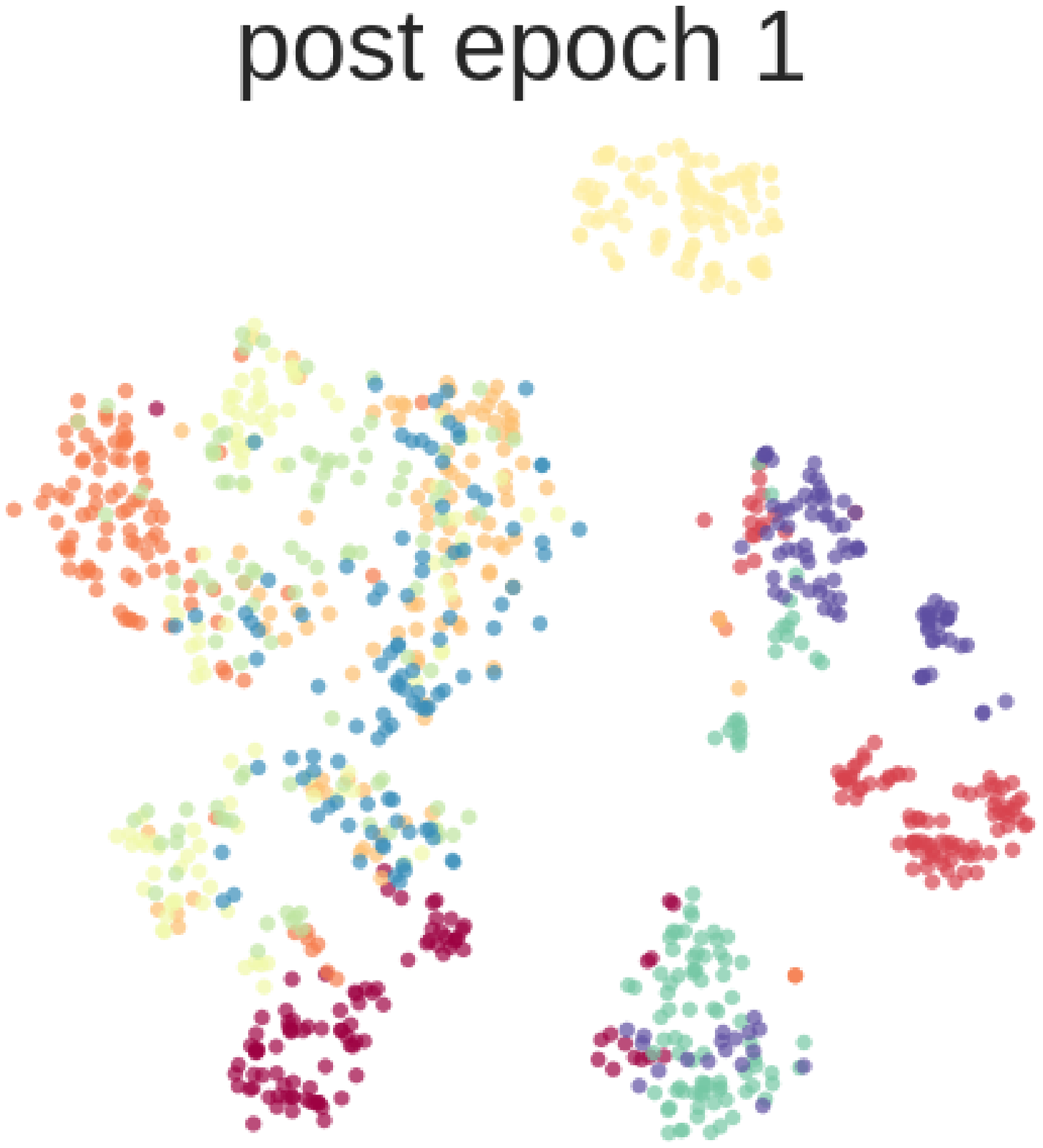} &
        \includegraphics[scale = 0.16]{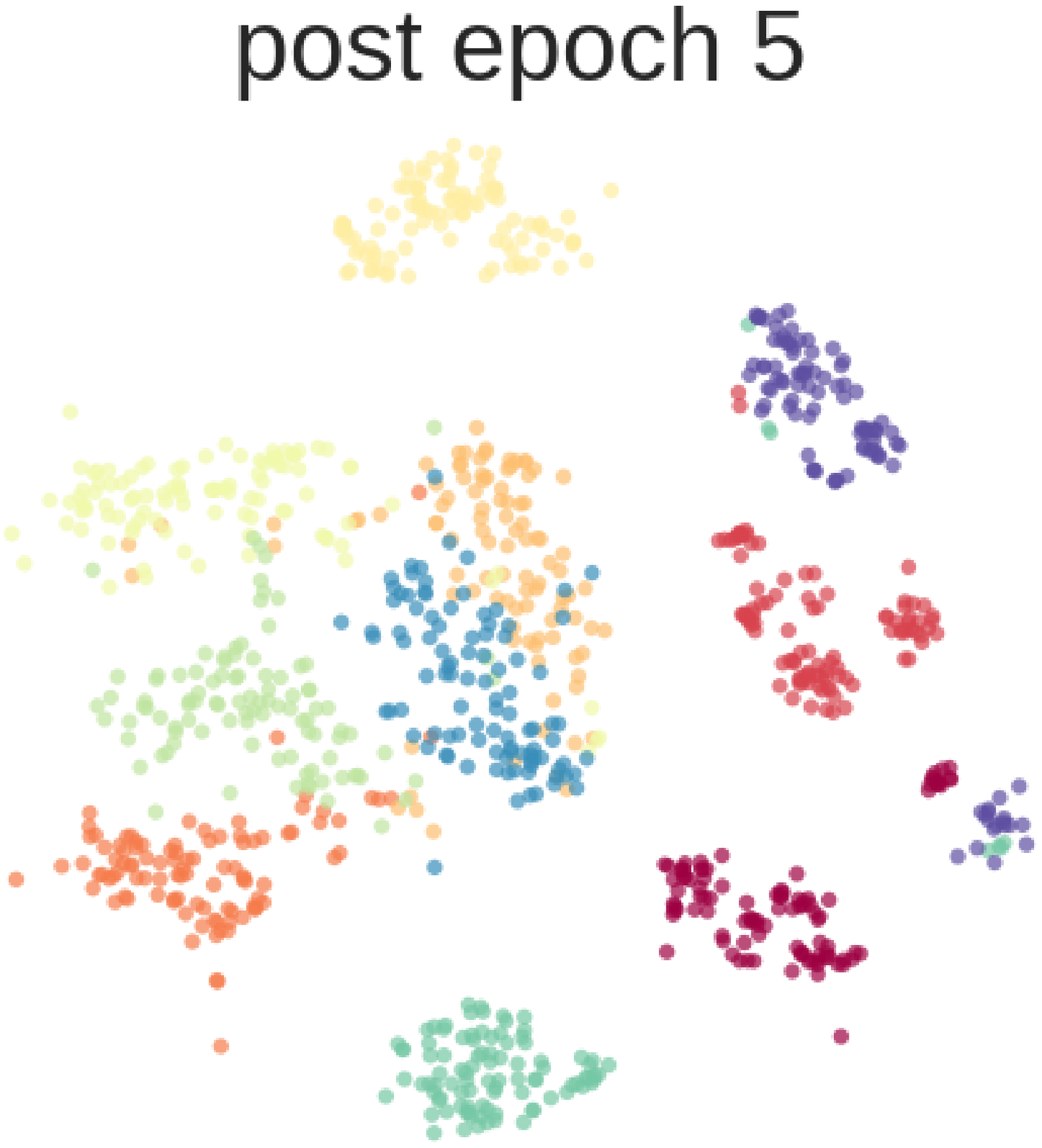} &
        \includegraphics[scale = 0.16]{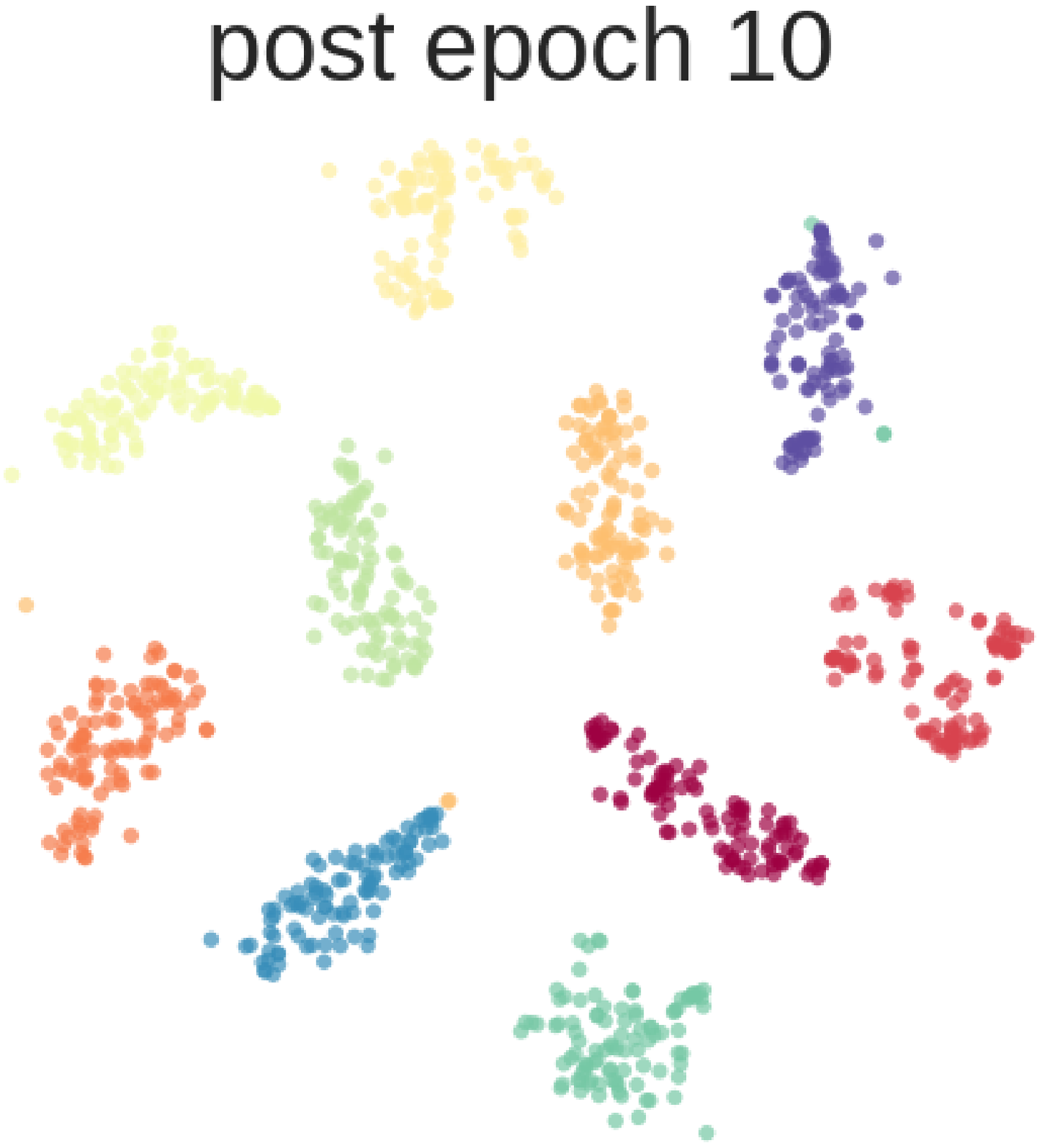} \\
	\end{tabular}
	\caption{Temporal transition of the post embeddings of 10 Twitter users.}
	\label{fig:embeddings}
\end{figure*}

Figure~\ref{fig:embeddings} shows the visualizations of the learned post embeddings of 10 Twitter users over the training epochs using t-SNE \citep{maaten2008visualizing}. We observe that over the training epochs, the post embeddings begin to cluster such that the posts of a user would move closer to one another in the embedding space while the posts of different users are ``pushed apart''. Consequently, when we aggregate the post embeddings of individual users to generate the user style embeddings, the embeddings between different users will also be pushed further apart in the embedding space over time. The temporal transitions of the post embeddings demonstrate \textsf{DeepSyle}'s capabilities to learn latent features which differentiate posts of different users and this important characteristic fundamentally enables us to apply \textsf{DeepStyles} to the AA problem.

%% file: introduction.tex
\textbf{Motivation.} Authorship attribution (AA), the task of finding the owner of a given text, plays a vital role in many applications. Particularly in the context of social media, AA is crucial in tackling the growing problem of online falsehood, vicious sock-puppets, and evidence gathering in forensic investigations. Although the AA task has been widely studied and many features have been explored \cite{layton2010authorship,koppel2014determining,sapkota2015not,schwartz2013authorship,sundararajan2018represents,sari2018topic}, the traditional document-based methods did not perform well on the social media posts as they tend to be shorter and less former \cite{koppel2014determining,shrestha2017convolutional}. 

To overcome the short-text challenge, several previous AA works have aggregated multiple social media posts of the same user into a single long document before training a classifier to predict the author labels using the long document features, \cite{ding2017learning,layton2010authorship}. Such an approach is, however, not effective in handling AA for individual query or a single social media post, which is necessary for some applications, e.g., detecting if two social media posts sharing a rumor originate from the same person. In a recent study, \cite{shrestha2017convolutional} used n-grams of individual Twitter posts to train a Convolutional Neural Network (CNN) model for AA. While CNN and other deep learning models can achieve improved accuracy for the AA task, they still have several limitations. Firstly, the existing models only utilized single input feature types (e.g., bi-gram) to represent the post. Such approaches neglect the rich types of features which had been explored in previous studies to overcome data sparsity in representing the user's writing styles from short text \cite{sari2018topic}. Secondly, the deep learning approach usually offers limited explainability on how some factors contribute to good (or bad) AA results as the models themselves are black boxes.

\textbf{Research Objectives.} In this paper, we address the limitations in existing methods and propose \textsf{DeepStyle}\footnote{Code implementation: https://gitlab.com/bottle\_shop/snlg/style/deepstyle}, an embedding-based framework specifically designed to learn the user style representations using deep learning. The key idea behind \textsf{DeepStyle} is to extract salient features from the users' social media posts. At the high level, our proposed \textsf{DeepStyle} is divided into two parts.  The first part learns a multi-view representation of a user's posts using deep neural networks over different types of features. Unlike existing deep learning AA methods which use Cross-Entropy loss as objective function, we adopt Triplet loss  \cite{cheng2016person} to learn the post embeddings such that posts that belong to the same user should be close to one another in the post embedding space, while posts from different users will be pushed further away from one another. The second part involves the use of an aggregation function to combine the post embeddings of a single user into the user's style embedding. The underlying intuition is that the style embeddings of different users should be far apart from one another in the embedding space, thereby distinguishing the users' writing styles. Finally, the learned user style embeddings are used to perform the AA task. 

There are benefits to learning user style embeddings. Firstly, the user style embedding allows us to perform AA in both unsupervised (i.e., finding the most similar user in the embedding space for a given query post) and supervised (i.e., as input to a classifier) settings. Secondly, the visualization of user style embedding enables us to gain an understanding of the feature differences and relationships between user writing styles. Lastly, beyond AA, the user style embedding can also be used for other forensic investigations such as clone \cite{xiao2015detecting} or sock-puppet account detection \cite{bu2013sock} by finding the nearest neighbors in the embedding space,i.e., user accounts who share most similar writing style.

\textbf{Contributions.}
Our main contributions in this work consist of the following.
\begin{itemize}
    \item We propose a novel embedding-based framework called \textsf{DeepStyle}, which utilizes multi-view representations of a user's post and Triplet loss objective functions to learn users' writing style for AA.
    \item Using \textsf{DeepStyle}, we manage to learn the latent representation of users' writing style. To the best of our knowledge, this is the first work that embeds user's writing style for AA. 
    \item We conduct extensive experiments on two real-world datasets from Twitter and Weibo. The experiment results show that \textsf{DeepStyle} consistently outperforms state-of-the-art methods regardless of the text's languages. 
\end{itemize}

%% file: related.tex
Authorship attribution (AA) is a widely studied research topic. Traditionally, feature engineering approaches were used to derive textual-related features from long documents such as emails and news articles. Subsequently, these derived features are used to train a classifier to identify an input document's owner \cite{koppel2009computational,stamatatos2009survey}. Commonly used features in AA includes character n-grams \cite{layton2010authorship,koppel2014determining,sapkota2015not,schwartz2013authorship}, lexical features, syntactical features, and document topics \cite{sundararajan2018represents,sari2018topic}. There are also recent studies that explore various deep learning approaches for AA. 
 
In recent years, the AA task has been studied in the social media context. A comprehensive survey \cite{rocha2016authorship} provides an overview of AA methods for social media content. The short social media content poses some challenges to the traditional AA methods, \cite{koppel2014determining,shrestha2017convolutional}. To overcome the short text challenge, the existing methods aggregate multiple social media posts of a user into a single document and before applying the AA methods, \cite{layton2010authorship,ding2017learning}. Such an approach cannot adequately handle queries at the single social media post level. Shrestha et al. \cite{shrestha2017convolutional} attempted to overcome this challenge by performing AA using a CNN model with character n-gram features. The n-gram CNN model considers each short text post of a user as input to train a CNN model for AA. In similar studies, Ruder et al. \cite{DBLP:journals/corr/RuderGB16c} investigated the use of CNN with various types of word and character level features for AA, while Boenningoff et al. \cite{Boenninghoff2019} proposed a hybrid CNN-LSTM on word level features to perform AA. Although these works achieved reasonably good accuracy for AA in social media, they had utilized single input feature types to represent the users' posts used in training the deep neural network classifier and offer limited explainability. In this work, we address the limitations of the state-of-the-art by proposing \textsf{DeepStyle}, an embedding-based framework that learns the user style embedding representations using multiple types of features derived from posts to perform AA.



%% file: model.tex



\subsection{Authorship Attribution Problem}
Let $U$ be a set of users in a social media platform, $U=\{u_1,u_2,...,u_N\}$, where $N$ is the total number of users in the platform. Each user $u$ has a set of posts, $P_{u}=\{p_{u,1},p_{u,1},...,p_{u,M_u}\}$, where $M_u$ is the number of posts belong to user $u$. Given $U$, $P_u$, and a query post $q$, we aim to predict the most likely author of $q$.

Traditionally, a classifier will be trained using the known posts of all users to predict the most plausible author of $q$. In this paper, we adopt a different approach by first learning post embeddings of a user $u$ with $M_u$ posts.  The symbol $e_{u,i}$ denotes the post representation of post $p_{u,i}$. The learned post embeddings $E_u = \{e_{u,1}, \cdots, e_{u,M_u}\}$ will then be aggregated to form the user style embedding $s_u$.  Finally, we compute for each user $u$ the cosine similarity between the query post embedding $e^q$ and $s_u$, denoted by $cos(e^q,s_u)$. Our model predicts the top similar user to be most plausible author of query post $q$.  In the following, we elaborate both the \textit{post embedding} and \textit{user style aggregation} modules.


\subsection{Post Embedding Module}
\begin{figure}[t]
    \centering
    \includegraphics[width=\textwidth]{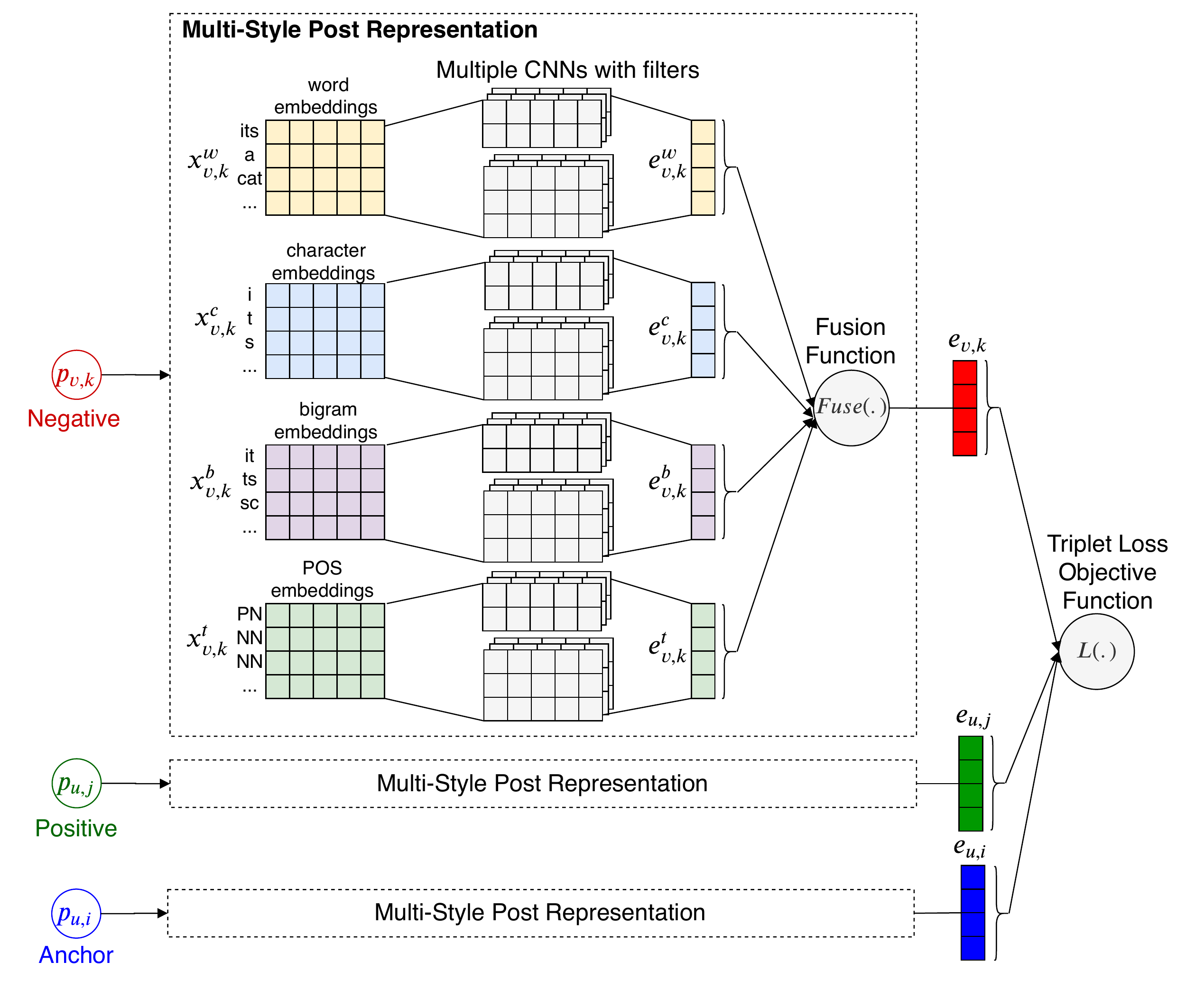}
    \caption{Post embedding module of the \textsf{DeepStyle} framework}
    \label{fig:framework_post_learning}
\end{figure}

Figure~\ref{fig:framework_post_learning} illustrates the overall architecture of the post embedding module in our \textsf{DeepStyle} framework. The module learns the post embedding representations using triplet loss. Each triplet is formed by first randomly sampling an anchor post $p_{u,i}$ and a positive post $p_{u,j}$ that belongs to the same user $u$. We also randomly sample a negative post $p_{v,k}$, which is a post belonging to another user $v$.  With a set of triplets of posts, we learn the post embeddings using \textit{Multi-Style Post Representation}. In this \textit{Multi-Style Post Representation} scheme, we first perform data pre-possessing to extract different types of feature embeddings (e.g., word, character, bigram, POS tags) for each post among the triplets. These feature embeddings will be used as input for a set of Convolution Neural Networks (CNNs) to process and output latent representations of the individual features. Subsequently, a fusion function is used to combine the latent representations and output a post embedding. Finally the three post embeddings, $e_{u,i}$, $e_{u,j}$, and $e_{v,k}$, will be optimized using the triplet loss objective function.       

\textbf{Feature Embeddings Extraction.} Previous studies have extensively explored the different types of features, such as words, characters, n-grams, syntactic, and semantic information for AA, \cite{stamatatos2009survey,ding2017learning,sari2018topic}. We leverage on the insights from previous studies to extract the following types of feature embeddings to represent a post $p_{u,i}$, $p_{u,j}$ or $p_{v,k}$. For simplicity, we use $p_{v,k}$ as an example.

For the four types of input embeddings $\{x^w_{v,k}, x^c_{v,k}, x^b_{v,k}, x^t_{v,k}\}$, we obtain four types of output latent representations through multi-level CNN layer , i.e., word latent representation $e^w_{v,k}$, character latent representation $e^c_{v,k}$, bi-gram latent representation $e^b_{v,k}$, and POS tag latent representation $e^t_{v,k}$.

\textbf{Post Features Fusion Function.} After learning the latent representations of the various types of feature embeddings, we use a fusion function to combine the various latent representations of a post:
\begin{equation}
    e_{v,k} = Fuse(e^w_{v,k}, e^c_{v,k}, e^b_{v,k}, e^t_{v,k})
\end{equation}
where $Fuse(\cdot)$ is a typical fusion kernel, and $e_{v,k}$ has the same dimension as the input latent representations. In our proposed framework, we explore five different implementations the of $Fuse(\cdot)$: \textit{Mean}, \textit{Max}, \textit{Multilayer perceptron (MLP)}, \textit{Attention}\cite{DBLP:conf/naacl/YangYDHSH16}, and \textit{Capsule} \cite{sabour2017dynamic}.

\subsection{User Style Aggregation Module}

A user publishes multiple posts on social media platforms. Each post contains some information that defines the user's writing style, which ultimately would help in authorship attribution. There are multiple ways to combine the post embeddings of a user $v$ to form the final user style embedding representation, $S_v$. In this paper, we introduce three types of aggregators for learning the post embedding of a user: \textit{Mean}, \textit{Max}, and \textit{Capsule} \cite{sabour2017dynamic}.

Using the above three aggregators, we aggregated the users' post embeddings to form the user style embeddings, $S=\{s_1,s_2,...,s_N\}$. The user style embeddings will be be used to perform AA. We will further evaluate the three aggregation methods in our experiments.

%% file: experiment.tex

\subsection{Experiment Setup}
\textbf{Datasets.} We conducted our experiments on two publicly available datasets, namely Weibo\footnote{https://hub.hku.hk/cris/dataset/dataset107483} and Twitter \cite{schwartz2013authorship}. The Twitter dataset is widely used in AA research \cite{schwartz2013authorship,shrestha2017convolutional}, while the Weibo dataset, which consists of short Chinese texts, demonstrates the language-agnostic nature of our model. The Twitter dataset contains tweets from 7,026 users, while the Weibo dataset contains posts from 9,819 users. In both datasets, each user has about 1000 posts.

\textbf{Baselines.} For evaluation, we compare our proposed \textsf{DeepStyle} model with the following state-of-the-art baselines:
\begin{itemize}
    \item \textbf{ToS}: Based on \cite{sari2018topic}'s work, ToS method derives content and style-based features from the users' posts to train a Logistic Regression classifier.  
    \item \textbf{CNN}: Following the studies in \cite{shrestha2017convolutional,DBLP:journals/corr/RuderGB16c}, we train a CNN model each on character features (called CNN-1), bigram features (called CNN-2), and word features ( called CNN-W).
    \item \textbf{LSTM}: Long-Short Term Memory (LSTM) model has also been implemented as a baseline in previous study \cite{shrestha2017convolutional} and the model has been  successfully used for text classification \cite{liu2016recurrent}. We train an LSTM at the character level (LSTM-1),  bigram level  (LSTM-2), and word (LSTM-W) level.
\end{itemize}
All the parameters of baselines are empirically set to yield optimal results. 




\subsection{Experiment Results}
Table \ref{tbl:resultspostK} shows the experiment results on Twitter and Weibo datasets. In this experiment, we use 100 posts from each user for training and test on 20 posts from each user. All users on Twitter and Weibo are used in this experiment. Note that the \textsf{DeepStyle} model in this experiment is configured with \textit{Capsule} as the feature fusion function to combine the various latent representations of a post and we select \textit{Mean} as the aggregator to combine the users' post embeddings into final user style embeddings. The same \textsf{DeepStyle} setting is used for the results reported in Tables~\ref{tbl:results} and \ref{tbl:resultspost}. From the results, we observe that \textsf{DeepStyle} outperforms the state-of-the-art baselines for different $k$ values using the $P@k$ measure. The improvements of \textsf{DeepStyle} over the best baselines (i.e., CNN-1 and CNN-2) are significant at 0.01 level using paired t-test. Similar to the previous study \cite{shrestha2017convolutional}, we observe that the CNN models outperform the LSTM models in both datasets. More interestingly, CNN-2 performs better than CNN-1 for the Twitter dataset, while CNN-1 outperforms CNN-2 for the Weibo dataset. This observation could be attributed to the difference in languages; Chinese characters may encode more semantics or definitive user styles. Hence, the Chinese character unigrams can yield better results.  \textsf{DeepStyle}, which utilizes multi-style post representations, outperforms all state-of-the-art baselines in both English (i.e., Twitter) and Chinese (i.e., Weibo) datasets.

\begin{table}[t]
\centering
\caption{P@k of various baselines and \textsf{DeepStyle} on Twitter and Weibo datasets. We vary $k = [1, 10, 50, 100]$. The improvement of \textsf{DeepStyle} over the best baselines (i.e., CNN-1) is significant at 0.01 level using paired t-test.}
\label{tbl:resultspostK}
\begin{tabular}{|l|cccc|cccc|}
\hline
 & \multicolumn{4}{c|}{\textbf{Twitter}}&\multicolumn{4}{c|}{\textbf{Weibo}}\\
{\textbf{Method}} & {\textbf{P@1}} & {\textbf{P@10}} & {\textbf{P@50}} & {\textbf{P@100}}& {\textbf{P@1}} & {\textbf{P@10}} & {\textbf{P@50}} & {\textbf{P@100}} \\
\hline \hline
ToS \cite{sari2018topic}  &9.8 &20.4 &32.2 &43.0 &12.1  &25.9  &38.6  &48.3  \\\hline
CNN-1 \cite{shrestha2017convolutional,DBLP:journals/corr/RuderGB16c} &18.7 &30.7 &43.1 &49.9 &22.0  &35.2  &48.4 &55.9  \\
CNN-2 \cite{shrestha2017convolutional} &20.1 &33.7 &47.6 &55.3 &17.5  &29.7  &41.3  &51.3  \\
CNN-W \cite{shrestha2017convolutional,DBLP:journals/corr/RuderGB16c} &16.4 &27.1 &37.6 &44.0 &21.0 &36.5  &48.6  &54.8  \\\hline
LSTM-1 &15.3 &25.6 &35.9 &42.0 &19.0  &33.6  &48.1 &55.5  \\
LSTM-2 \cite{shrestha2017convolutional} &15.5 &28.0 &40.9 &48.6 &15.8 &31.5  &43.5  &49.8  \\
LSTM-W &11.7 &23.7 &41.2 &48.1 &18.7  &34.1  &48.0  &55.2  \\\hline
DeepStyle &\textbf{21.4} &\textbf{37.1} &\textbf{51.5} &\textbf{59.2} & \textbf{23.9} & \textbf{38.1} & \textbf{51.6} &\textbf{58.9} \\
\hline
\end{tabular}
\end{table}

\textbf{Varying number of users and posts.} Similar to previous studies \cite{shrestha2017convolutional,schwartz2013authorship}, we also want to explore how \textsf{DeepStyle} performs when the AA problem becomes more difficult, i.e., when the number of users increases or when the number of tweets per author is reduced. Table~\ref{tbl:results} shows the $P@1$ of DeepStyle and various baselines varying the number of users in Twitter and Weibo from 50 to all users. For this experiment, we use 100 posts from each user for training and test on 20 posts from each user. We note that although the precision decreases with the increasing number of users, \textsf{DeepStyle} is able to outperform state-of-the-art baselines consistently. Even in the worse case of all users, i.e., over 7,000 and 9,000 Twitter and Weibo users respectively, \textsf{DeepStyle} is still able to outperform the best baselines and the improvements over the best baselines are significant at 0.01 level using paired t-test.

Besides varying the number of users, we also conducted experiments to evaluate the models' performance with different amounts of training data. Table~\ref{tbl:resultspost} shows the results of the models trained with a different number of posts per user for 50 randomly selected Twitter and Weibo users. We observe the $P@1$ of various models drops as we reduce the number of posts. However, \textsf{DeepStyle} still consistently outperforms the baselines even with reduced training data.

\begin{table*}[t]
\centering
\caption{P@1 for \textsf{DeepStyle} and baselines on varying number of users in Twitter and Weibo. }
\label{tbl:results}
\begin{tabular}{|l|ccccc|ccccc|}
\hline
 & \multicolumn{5}{c|}{\textbf{\#Twitter Users}} & \multicolumn{5}{c|}{\textbf{\#Weibo Users}} \\
{\textbf{Method}} & {\textbf{50}} & {\textbf{100}} & {\textbf{500}} & {\textbf{5000}} & {\textbf{All}} & {\textbf{50}} & {\textbf{100}} & {\textbf{500}} & {\textbf{5000}} & {\textbf{All}} \\\hline \hline
ToS \cite{sari2018topic} &48.3 &31.2 &22.0 &15.7 &9.8 &50.5 &39.5 &22.2 &16.0 &12.1 \\\hline
CNN-1 \cite{shrestha2017convolutional,DBLP:journals/corr/RuderGB16c} &60.9 &47.1 &35.6 &23.1 &18.7 &61.3 &52.6 &38.8 &25.1 &22.0 \\
CNN-2 \cite{shrestha2017convolutional} &59.3 &46.7 &36.5 &24.6 &20.1 &47.6 &42.4 &30.4 &18.7 &17.5 \\
CNN-W \cite{shrestha2017convolutional,DBLP:journals/corr/RuderGB16c} &50.0 &37.1 &28.4 &17.7 &16.4 &56.8 &48.3 &36.2 &23.8 &21.0 \\\hline
LSTM-1 &40.3 &29.6 &27.3 &16.1 &15.3 &37.5 &32.5 &30.9 &20.1 &19.0 \\
LSTM-2 \cite{shrestha2017convolutional} &45.1 &34.5 &25.2 &16.5 &15.5 &28.1 &26.4 &19.6 &16.8 &15.8 \\
LSTM-W &35.0 &22.8 &19.3 &12.0 &11.7 &35.0 &33.4 &28.8 &19.8 &18.7 \\\hline 
DeepStyle & \textbf{64.8} & \textbf{51.2} & \textbf{37.9} & \textbf{25.5} & \textbf{21.4} & \textbf{65.2} & \textbf{56.6} & \textbf{42.8} &\textbf{27.9} &\textbf{23.9} \\
\hline
\end{tabular}
\end{table*}
 

\begin{table}[t]
\centering
\caption{P@1 for \textsf{DeepStyle} and baselines on varying number of posts for 50 Twitter and Weibo users.}
\label{tbl:resultspost}
\begin{tabular}{|l|cccc|cccc|}
\hline
 & \multicolumn{4}{c|}{\textbf{Twitter\#posts}}&\multicolumn{4}{c|}{\textbf{Weibo\#posts}}\\
{\textbf{Method}} & {\textbf{500}} & {\textbf{200}} & {\textbf{100}} & {\textbf{50}}& {\textbf{500}} & {\textbf{200}} & {\textbf{100}} & {\textbf{50}} \\
\hline \hline
ToS \cite{sari2018topic} &47.4 &45.1 &40.5 &36.2 & 49.2 & 44.9 & 40.8 & 34.7 \\\hline
CNN-1 \cite{shrestha2017convolutional,DBLP:journals/corr/RuderGB16c} &59.6 &56.0 &51.7 &49.3 & 60.5 & 56.7 & 52.2 & 49.1 \\
CNN-2 \cite{shrestha2017convolutional} &61.7 &54.1 &48.4 &44.4 & 59.4 & 56.3 & 51.8 & 47.5 \\
CNN-W \cite{shrestha2017convolutional,DBLP:journals/corr/RuderGB16c} &52.1 &48.0 &43.3 &40.2 & 58.0 & 53.1 & 47.9 & 44.2 \\\hline
LSTM-1 &55.7 &49.5 &39.8 &32.9 &54.7 &48.7 &37.6  &32.0  \\
LSTM-2 \cite{shrestha2017convolutional} &53.2 &39.4 &34.2 &30.4 &51.0  &42.4 &35.5 &31.4  \\
LSTM-W &47.6 &41.5 &35.2 &30.5 &52.8  &41.9  &36.1  &31.2  \\\hline
DeepStyle &\textbf{62.8} &\textbf{58.7} &\textbf{56.3} &\textbf{50.9} & \textbf{62.5} & \textbf{58.2} & \textbf{56.1} &\textbf{51.1} \\
\hline
\end{tabular}
\end{table}

%% file: conclusion.tex
In this paper, we proposed a novel embedding-based framework call \textsf{DeepStyle}, which utilized multiple feature types to learns the embedding representation of the user's writing style for authorship attribution (AA). We evaluated \textsf{DeepStyle} using publicly available real-world datasets, and our extensive experiments have shown that \textsf{DeepStyle} is language-agnostic and outperforms the baselines in different experimental settings. For future works, we will like to consider the evolution of the user's writing style by modeling the time-series aspect of the user's posts and consider other better forms of visualizations to explain the user's writing style. 
